\pdfoutput=1

\documentclass[11pt,table,dvipsnames]{article}

\usepackage[final]{acl}
\usepackage{times}
\usepackage{latexsym}
\usepackage[utf8]{inputenc}

\usepackage[T1]{fontenc}

\usepackage[utf8]{inputenc}

\usepackage{microtype}

\usepackage{inconsolata}
\usepackage{microtype}
\usepackage{latexsym}
\usepackage{dsfont}
\usepackage{booktabs} 
\usepackage{CJKutf8}
\usepackage{graphicx}
\usepackage{bigstrut}
\usepackage{multirow}
\usepackage{amsmath}
\usepackage{multirow, makecell, caption}
\usepackage{floatrow}
\newfloatcommand{capbtabbox}{table}[][\FBwidth]
\usepackage{xcolor}
\usepackage[normalem]{ulem}
\usepackage{amssymb}
\usepackage{amsfonts}
\usepackage{multicol}
\usepackage{tipa}
\usepackage{amsmath}
\usepackage{bm}
\usepackage[ruled,linesnumbered]{algorithm2e}
\usepackage{tikz}
\usepackage{paralist}
\usepackage{IEEEtrantools}
\usepackage[switch]{lineno}
\usepackage{enumitem}
\usepackage{multirow, makecell, caption}
\usepackage{arydshln}
\usepackage{verbatim}
\usepackage[normalem]{ulem}
\usepackage{amssymb}
\usepackage{amsfonts}
\usepackage{multicol}
\usepackage{amssymb}
\usepackage{pifont}
\usepackage{csquotes}
\usepackage{xspace}
\usepackage{paralist}
\usepackage{mdwlist}
\usepackage{subcaption}
\usepackage{makecell}
\usepackage{array}
\usepackage{bm}
\usepackage{colortbl}
\usepackage{dsfont}
\usepackage{url}
\usepackage{booktabs} 
\usepackage{graphicx}
\usepackage{tabularx}
\usepackage{amsmath}
\usepackage{bbm}
\usepackage{pgfplots}
\usepackage{soul} 
\usepackage{listings}
\usepackage[tableposition=top]{caption}
\usepackage{color,xcolor} 

\definecolor{Ground}{RGB}{255,184,55}
\definecolor{Rice}{RGB}{251,248,238}
\definecolor{Dirt}{RGB}{191,169,115}
\definecolor{Pink}{RGB}{226,184,176}
\definecolor{Violet}{RGB}{163,148,170}
\definecolor{mygray}{RGB}{226, 226, 226}

\newcommand{\method}{\textsc{MIRROR}\xspace}
\newcommand{\bench}{\textsc{RoleThink}\xspace}

\newcolumntype{g}{>{\columncolor{Ground!10}}c}
\newcolumntype{d}{>{\columncolor{Dirt!10}}c}
\newcolumntype{f}{>{\columncolor{Pink!10}}c}
\newcolumntype{v}{>{\columncolor{Violet!10}}c}

\makeatletter
\def\adl@drawiv#1#2#3{%
        \hskip.5\tabcolsep
        \xleaders#3{#2.5\@tempdimb #1{1}#2.5\@tempdimb}%
                #2\z@ plus1fil minus1fil\relax
        \hskip.5\tabcolsep}
\newcommand{\cdashlinelr}[1]{%
  \noalign{\vskip\aboverulesep
           \global\let\@dashdrawstore\adl@draw
           \global\let\adl@draw\adl@drawiv}
  \cdashline{#1}
  \noalign{\global\let\adl@draw\@dashdrawstore
           \vskip\belowrulesep}}
\makeatother

\title{\textit{Guess What I am Thinking:} A Benchmark for Inner Thought Reasoning of Role-Playing Language Agents}

\author{
    Rui Xu\textsuperscript{\rm $\heartsuit\spadesuit$}\thanks{~~Work is done during internship at INF.}, 
    MingYu Wang\textsuperscript{\rm $\heartsuit$},
    XinTao Wang\textsuperscript{\rm $\heartsuit$}\\
    \bf Dakuan Lu\textsuperscript{\rm $\spadesuit$}\thanks{~~Corresponding authors.},
    Xiaoyu Tan\textsuperscript{\rm $\spadesuit$}\footnotemark[2],
    Wei Chu\textsuperscript{\rm $\spadesuit$},
    Yinghui Xu\textsuperscript{\rm $\heartsuit\spadesuit$}\footnotemark[2]\\
    \textsuperscript{\rm $\heartsuit$}Fudan University
    \textsuperscript{\rm $\spadesuit$}INF Technology (Shanghai) Co., Ltd.
    \\
    \texttt{rxu24@m.fudan.edu.cn},
    \texttt{xuyinghui@fudan.edu.cn}
}

\begin{document}
\maketitle
\begin{abstract}
Recent advances in LLM-based role-playing language agents (RPLAs) have attracted broad attention in various applications. While chain-of-thought reasoning has shown importance in many tasks for LLMs, the internal thinking processes of RPLAs remain unexplored. Understanding characters’ inner thoughts is crucial for developing advanced RPLAs. In this paper, we introduce \bench, a novel benchmark constructed from literature for evaluating character thought generation. We propose the task of inner thought reasoning, which includes two sets: the gold set that compares generated thoughts with original character monologues, and the silver set that uses expert-synthesized character analyses as references. To address this challenge, we propose \method, a chain-of-thought approach that generates character thoughts by retrieving memories, predicting character reactions, and synthesizing motivations. Through extensive experiments, we demonstrate the importance of inner thought reasoning for RPLAs, and \method consistently outperforms existing methods. Resources are available at \url{https://github.com/airaer1998/RPA_Thought}.

\begin{figure}[!ht]
    \centering
    \includegraphics[width=0.95\linewidth]{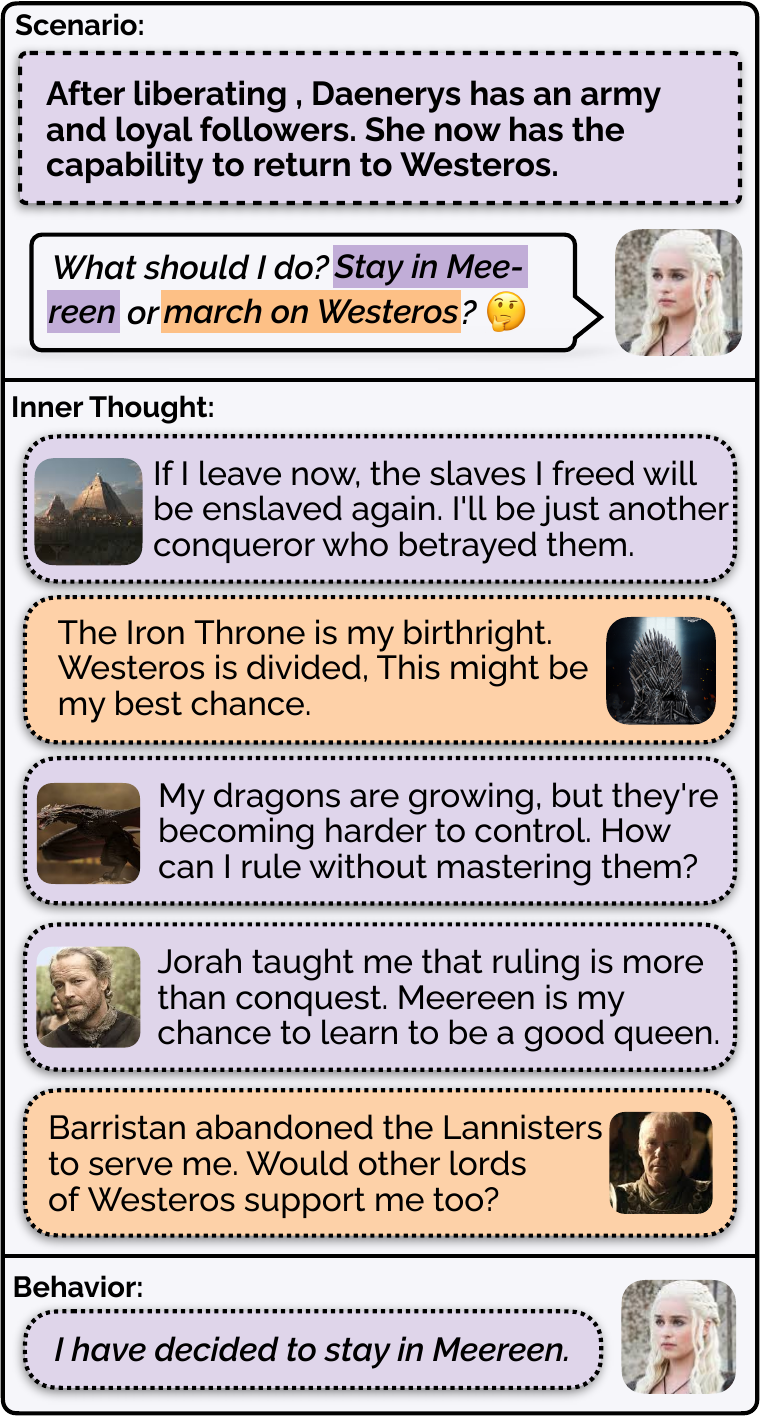}
    \caption{The process of characters performing inner thought reasoning. When facing specific scenarios, characters generate different inner thoughts, which reflect a deep understanding of the character while influencing their behavior.}
    \label{fig:front}
\end{figure}
\end{abstract}

\section{Introduction}

Recent advances in large language models (LLMs) have enabled the development of role playing language agents (RPLAs)~\citep{shao2023character, chen2024persona}, which are now widely used in applications from character chatbots~\citep{xu2024mindecho} to game NPCs~\citep{wang2023voyager,ran2024capturing}. 
While chain-of-thought reasoning~\citep{wei2023chainofthoughtpromptingelicitsreasoning} has shown significant success in various LLM tasks~\citep{chu2024navigateenigmaticlabyrinthsurvey,xiang20252reasoningllmslearning}, the inner thought process of RPLAs remains unexplored.
To address this gap, research on generating high-quality character thought data both helps us understand character motivations and shows strong potential for improving role-playing capabilities.

Prior research has explored LLMs' ability to understand characters in fictional works, primarily focusing on basic tasks such as character prediction~\citep{brahman2021let, yu2022few} and personality prediction~\citep{yu2023personality}, lacking deep analysis of character behaviors.
Recently, the research focus has shifted to role-playing, where LLMs have demonstrated strong performance in basic role-playing tasks, such as knowledge replication~\citep{zhou2023characterglm} and speaking style imitation~\citep{li2023chatharuhi}.
However, these models show limited capabilities in complex tasks involving decision-making~\citep{xu2024character} and psychological reasoning~\citep{wang2023does}.
As shown in Figure \ref{fig:front}, generating inner thought processes before actions enables both in-depth analysis of character motivations and better completion of subsequent behaviors. However, constructing high-quality character inner thought data remains a significant challenge.


In this paper, we systematically evaluate LLMs' ability to generate thought chains for role-playing models. 
We construct \bench Benchmark using \textit{A Song of Ice and Fire} as the source material, which contains extensive character monologues and rich character behavior analyses from fan websites.
We propose the task of inner thought reasoning. In this task, given a character profile, LLMs are required to generate plausible thoughts based on the current scenario, which are then evaluated against reference thoughts. Based on different reference types, we divide the task into two sets: Gold and Silver.
The Gold Set uses character monologues from the original novels as references, while the Silver Set employs synthesized data from fan websites and literary experts' character analyses. 
To ensure comprehensive evaluation, we employ both automatic metrics and human assessment methods.

To better analyze and address this task, we conduct extensive experiments with various LLMs. 
Based on our analysis, we propose \textbf{M}emory \textbf{I}ntegration and \textbf{R}ole \textbf{R}easoning with \textbf{O}bservation  \textbf{R}eflection (\method), a chain-of-thought approach that retrieves multiple relevant memories from the current scenario, predicts reactions of related characters and environments, and summarizes them into character motivations. 
Models enhanced with \method consistently achieve higher scores on \bench. 
To further validate the benefits of reasoning processes in RPLAs, we evaluate on multiple role-playing benchmarks. 
The results demonstrate that enabling RPLAs to think before acting improves performance across various downstream tasks.

Our contributions are summarized as follows:
\begin{inparaenum}[\it 1)]
\item We present the first study on the inner thought process behind RPLAs and construct the first benchmark for evaluating this process.
\item We propose \method, a method that better generates character inner thought processes.
\item We conduct extensive experiments with different LLMs, validating the importance of inner thought processes across various role-playing downstream tasks.
\end{inparaenum}

\section{Related Work}
\label{sec:related}

\subsection{Character Understanding}
Understanding characters is crucial for natural language processing systems to comprehend narrative texts.
Recent research has explored various aspects of character analysis, including personality traits~\citep{yu2023personality}, relationship networks~\citep{chen2016character}, and behavioral patterns~\citep{brahman2021let}.
~\citet{yuan2024evaluating} attempt to use LLMs to summarize character profiles from source materials.
Our research extends this by exploring the generation and analysis of characters' inner thoughts, providing deeper insights into their decision-making processes.
Previous work has developed benchmarks for character identification~\cite{sang2022tvshowguess} and question answering~\cite{kovcisky2018narrativeqa}, but these primarily evaluate surface-level text understanding.
In contrast, we propose methods to model characters' internal reasoning processes, including memory recall and theory of mind thinking.

\subsection{Role Playing Language Model}
Recent advances in LLMs have enabled the development of more sophisticated character role-playing systems~\citep{wang2025cosercoordinatingllmbasedpersona}.
Current research in RPLAs focuses on generating character-consistent responses in dialogues, using methods such as character dialogue fine-tuning~\cite{li2023chatharuhi,wang2023rolellm}, memory retrieval~\cite{shao-etal-2023-character,chen2023large}, and evaluation metrics~\cite{tu2024charactereval,zhou2023characterglm} to ensure response consistency with character tone, personality, and knowledge.
While these approaches improve surface-level interaction quality, they do not address characters' underlying psychological processes.
\citet{xu2024character} propose evaluating role-playing ability through characters' decisions from original works, which requires deeper character understanding.
\citet{wang2023does} further explore this direction by evaluating characters through psychological interviews and personality fidelity, showing that current models often lack consistent inner reasoning processes.
Similar findings are reported in other evaluation frameworks~\cite{chen2024persona,shen2023roleeval,yuan2024evaluating}.
Our work addresses these limitations by introducing explicit thought generation in character role-playing.
We model how characters recall relevant memories, consider others' perspectives, and reason about their decisions.
This approach provides a more complete framework for understanding and reproducing character behaviors.

\section{\bench Benchmark}
\label{sec:bench}

\subsection{Task Formulation}
Character Inner Thought Reasoning aims to generate characters' thinking processes in specific scenarios. Given a character profile $P$ and a scenario description $S$, the task requires LLMs to generate the thoughts $T$ that led to their behavior. 

We divide our benchmark into two sets based on different sources of reference. The Gold Set aims to recover a character's original thoughts from the novels. Given a character profile $P$ and a scenario description $S$, where the character's original thoughts are masked, the model generates thoughts $T$ that are evaluated against $T_{gold}$, which is directly extracted from the character's thoughts in the source material. The Silver Set focuses on generating plausible thoughts for given scenarios. With the same input format of profile $P$ and scenario $S$, the model generates thoughts $T$ that is evaluated against $T_{silver}$ collected from literary experts and fan communities' analyses of the character's inner thoughts in that scenario.

\subsection{Gold Set}
\begin{table}[!t]
  \centering
  \small
    \begin{tabularx}{\linewidth}{X}
    \toprule
\rowcolor[gray]{0.95} \multicolumn{1}{c}{\textbf{Gold Set}}\\
\textbf{Character}: Ned Stark\\
\textbf{Scenario}:
"Pain is a gift from the gods, Lord Eddard," Grand Maester Pycelle told him... Cersei faced him calmly, without flinching. "He saw us together. You love your children, don't you?" Ned thought: [MASK]\\
\textbf{Reference}:
\textcolor{teal}{
What would he do if another child threatened the lives of Robb, Sansa, Arya, Bran, or Rickon? Or even, what would Catelyn do if Jon threatened the lives of her own children? He didn't know, and he prayed he would never have to find out.}\\
\hdashline  
\textbf{Character}: Stannis Baratheon\\
\textbf{Scenario}: The smoke from the burning gods darkened the morning sky... Stannis thought, [MASK]\\
\textbf{Reference}: \textcolor{teal}{I begged them humbly, and all I got was mockery. I'll never be so weak again, and no one will ever mock me again. The Iron Throne is mine by right, but how do I take it? There are four kings in the realm, and the other three all have more gold and men than I do. All I have are ships... and her - the Red Woman.}\\
\midrule
\rowcolor[gray]{0.95} \multicolumn{1}{c}{\textbf{Silver Set}}\\
\textbf{Character}: Tyrion Lannister \\
\textbf{Scenario}: Heavy footsteps sounded outside the wooden door, and Tyrion Lannister knew his time had come... As expected, he found his father in the small tower that served as a privy. Lord Tywin had his robe gathered around his hips, and he looked up at the sound of footsteps. \\
\textbf{Reference}: \textcolor{brown}{Standing before him with a crossbow in my hands, I stare at Tywin Lannister. His composure and arrogance remain unchanged, while my heart churns with every painful memory from the past.}\\
\textcolor{brown}{I remember Tysha, and how father mercilessly crushed our love. It was the only true love I ever felt, yet he tore it into a wound that would never heal.}\\
\textcolor{brown}{For years, I've been nothing but a "monster" in his eyes, a stain on the family. No matter how hard I tried to prove myself, I never had the dignity of a son in his eyes.}\\
\textcolor{brown}{Shae's betrayal made me realize I truly had nothing in this world. And father's insult to Tysha was the final straw I could not forgive.}\\
\textcolor{brown}{Even if I choose to lower the crossbow now, I know father would never spare me. In his eyes, I am already a dead man. He would find a "proper" way to eliminate this family stain, just as he handles all other threats.}\\
\textcolor{brown}{Looking at this eternally proud man before me, I finally understand that only by casting away Tywin Lannister's shadow can I find true freedom.}\\
\textcolor{brown}{And so, I make my decision.} \\
    \bottomrule
    \end{tabularx}
  \caption{Case study of \bench, including data from two gold sets and one silver set. }
  \label{tab:case}
\end{table}

The Gold Set requires LLMs to recreate characters' thoughts from the original book. To collect high-quality thought data, we input each POV chapter ($\sim$10k tokens) into GPT-4o to identify key characters and detect their inner thoughts in these sections. The detected thought segments are then manually filtered to ensure quality. Table \ref{tab:case} shows two complete examples from the Gold Set. 
We mask the character thoughts in the chapters and use the text before each mask as scenario data. LLMs need to use character profiles to generate the masked thought content.

\subsection{Silver Set}
The Silver Set requires LLMs to generate character thoughts not present in the original book. 
To collect reference data, we gather character analysis from three fan websites\footnote{\url{https://gameofthrones.fandom.com/wiki/Wiki_of_Westeros}}\textsuperscript{,}\footnote{\url{https://asoiaf.huijiwiki.com/wiki/}}\textsuperscript{,}\footnote{\url{https://www.facebook.com/IceandFireCH/}}. We use GPT-4o to process this data, summarizing character motivations and locating specific thought points in the chapters. 
The collected data is then manually filtered to ensure quality. 
Table \ref{tab:case} shows a complete example from the Silver Set, where the located chapter is split according to the thought occurrence point. 
The content before the split serves as the scenario, and LLMs need to use character profiles to generate possible thinking processes at this point. 
The prompts and manual filtering criteria for both sets are provided in Appendix \ref{sec:appendix_prompt}.

\begin{figure*}[ht]
    \centering
    \includegraphics[width=\linewidth]{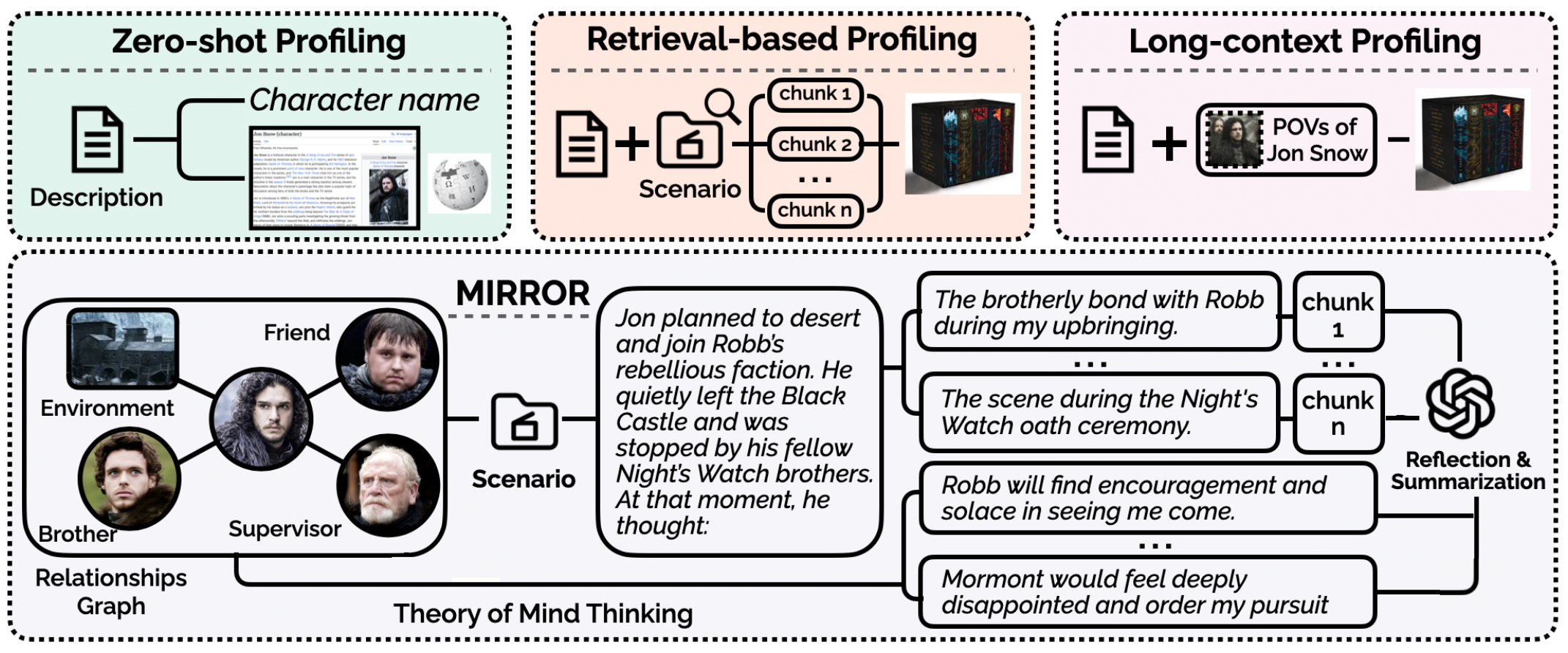}
    \caption{The framework of \method and its comparison with other profiling methods. Given a scenario, \method generates potential memories and objects, retrieves relevant memories, predicts object reactions, and synthesizes the results. Both scenarios and outputs are represented as summarized text for clarity.}
    \label{fig:data}
\end{figure*}

\subsection{Dataset}
\label{sec:bench_dataset}

We build our dataset using \textit{A Song of Ice and Fire} series, which is particularly suitable for our task due to its point-of-view narrative structure, abundant character monologues, and rich character analyses from fan communities and literary experts. For the Gold Set, we collect 405 data points from 22 characters, with an average scenario length of 5,036 tokens and an average reference length of 90 tokens. For the Silver Set, we collect 211 data points from 18 characters, with an average scenario length of 7,801 tokens and an average reference length of 301 tokens.

\subsection{Evaluation}

We employ three evaluation approaches to evaluate model performance on both sets comprehensively. (1) First, we use \textit{automatic text evaluation metrics} including BLEU~\citep{papineni2002bleu} and ROUGE-L~\citep{lin2004rouge}. However, we observe that in the Gold Set, the original thought data ($T_{gold}$) is often accurate but incomplete, limiting these metrics' effectiveness.
(2) To address this limitation, we introduce \textit{model-based automatic evaluation methods}. We use both discriminative NLI models and generative LLMs. For NLI evaluation, we employ mDeBERTa-v3-base-xnli\footnote{\url{https://huggingface.co/MoritzLaurer/mDeBERTa-v3-base-xnli-multilingual-nli-2mil7}}, a NLI model trained on DeBERTa~\citep{he2021debertadecodingenhancedbertdisentangled}. This model classifies text pairs into entailment, contradiction, or neutral relationships, and we use the entailment probability as the evaluation score. 
For LLM evaluation, we use GPT-4o to rate the coverage of generated thoughts against the reference on a 1-5 scale.
(3) Finally, we conduct \textit{human evaluation} with five crowdsourced annotators using the same scoring criteria as the LLM evaluation. 
Detailed evaluation criteria are provided in Appendix \ref{sec:appendix_eval}.

\section{\method}
\label{sec:method}
To generate character inner thoughts, we need to address two key challenges: locating relevant key evidence from the character's long-context memory, and reasoning about possible reactions of related objects from the character's perspective.
We propose Memory Integration and Role Reasoning with Observation Reflection (\method), a chain-of-thought approach with three steps: memory recall, Theory of Mind thinking, and reflection \& summarization. We describe each step in detail below.

\subsection{Memory Recall}
\label{sec:memory_recall}
Retrieving relevant memories is critical for generating reasoning chains. Previous methods~\citep{li2023chatharuhi,xu2024mindecho} retrieve memories based on current scenarios, but these often lack direct semantic connections and require complex reasoning. Other approaches~\citep{xu2024character,yuan2024evaluating} use long-context memory as prompts, but models struggle to process such lengthy inputs, limiting their performance.
As shown in Figure \ref{fig:data}, \method guides RPLAs to first recall related events based on the current scenario, then retrieve memories for each event. On average, each scenario triggers 2.6 related events. We split character memories into 1k token chunks. For each event, we compute relevance scores using cosine similarity between event and chunk embeddings from OpenAI's text-embedding-ada-002 model~\citep{neelakantan2022text}, and retrieve the chunks with the highest scores.

\subsection{Theory of Mind Thinking}
Theory of Mind~\citep{Premack_Woodruff_1978,street2024llms} refers to our ability to model others' mental states during communication and predict their responses to adjust our outputs. This concept has been applied to enhance social reasoning in LLMs~\citep{zhou2023sotopia}.
Unlike traditional reasoning tasks where models primarily focus on logical deduction, character-centric reasoning requires a deep understanding of social dynamics and interpersonal relationships. In \method, we guide characters to analyze and predict reactions of related objects (including characters, groups, and environments). This process consists of two steps: predicting potential objects in the scenario and analyzing possible responses for each object. By explicitly modeling others' perspectives, our approach helps characters make more socially aware and contextually appropriate decisions. As shown in Figure~\ref{fig:data}, when Jon Snow considers deserting the Night's Watch to join Robb's rebellion, the objects that might react include his brother Robb, his supervisor Lord Mormont, his fellow Night's Watch brothers, and the environment of the Night's Watch itself. The model predicts these objects' reactions - for instance, Robb might be encouraged by this, but Lord Mormont would be disappointed, while the Night's Watch would pursue any deserters.

\subsection{Reflection \& Summarization}
After obtaining memory chunks and Theory of Mind thinking results, we guide the model to organize all information. The model first filters out irrelevant content through reflection, then summarizes the remaining information to generate the final character inner thought process.
This two-stage process is crucial for maintaining the coherence and relevance of character thoughts. The reflection stage helps eliminate noise and tangential information that might distract from the core decision-making process. 
The summarization stage then synthesizes the filtered information into a coherent thought process that aligns with the character's established personality and motivations. This ensures that the generated inner thoughts are not merely a collection of related information, but a structured reasoning process that reflects the character's unique perspective and decision-making style.
All prompts related to \method can be found in the Appendix~\ref{sec:appendix_mirror}.

\section{Experiment Settings}
\label{sec:experiment}

\subsection{Character Profiling}
A character's profile refers to the input prompt for the role-playing model. As shown in Figure~\ref{fig:data}, in addition to \method, we employ three baseline methods to construct character profiles:
(1) Zero-shot profiling: Provides LLMs with only the character's name and a brief introduction from Wikipedia ($\sim$200 tokens).
(2) Retrieval-based profiling: Besides the character's name and introduction, uses the same settings as \method in Section \ref{sec:memory_recall}, retrieves the three most relevant memory chunks based on the current scenario, with each chunk being 1k tokens in length. These memories are included as part of the profile.
(3) Long-context profiling: Uses all POV data of the target character before the scenario as the character's profile, with an average length of 85k tokens and a maximum length of 381k tokens.

\newcolumntype{a}{>{\columncolor{BlueGreen!10}}c}
\newcolumntype{b}{>{\columncolor{brown!10}}c}
\newcolumntype{d}{>{\columncolor{Violet!10}}c}
\newcolumntype{q}{>{\columncolor{Green!10}}r}
\begin{table*}[t]
\setlength{\tabcolsep}{4pt}
\centering  
\small
\begin{tabular}{llbbbbbddddd}
\toprule
\multicolumn{1}{c}{\multirow{3}[1]{*}{\makecell{\textbf{Profiling}\\\textbf{Method}}}} &  
\multicolumn{1}{c}{\multirow{3}[1]{*}{\textbf{Base Model}}} &
\multicolumn{5}{c}{\textbf{Gold Set}} & 
\multicolumn{5}{c}{\textbf{Silver Set}}\\
\cmidrule(lr){3-7} \cmidrule(lr){8-12}
& & \multicolumn{2}{c}{\textbf{Text Metrics}} & \multicolumn{2}{c}{\textbf{Model Eval}} & \multicolumn{1}{c}{\textbf{Human}} & \multicolumn{2}{c}{\textbf{Text Metrics}} & \multicolumn{2}{c}{\textbf{Model Eval}} & \multicolumn{1}{c}{\textbf{Human}}\\
\cmidrule(lr){3-4} \cmidrule(lr){5-6} \cmidrule(lr){7-7} \cmidrule(lr){8-9} \cmidrule(lr){10-11} \cmidrule(lr){12-12}
& & \multicolumn{1}{c}\textbf{BLEU} & \multicolumn{1}{c}\textbf{ROUGE-L} & \multicolumn{1}{c}\textbf{NLI} & \multicolumn{1}{c}\textbf{LLM} & \multicolumn{1}{c}\textbf{Human} & \multicolumn{1}{c}\textbf{BLEU} & \multicolumn{1}{c}\textbf{ROUGE-L} & \multicolumn{1}{c}\textbf{NLI} & \multicolumn{1}{c}\textbf{LLM} & \multicolumn{1}{c}\textbf{Human}\\
\midrule
\addlinespace[0.05cm]
\multirow{5}{*}{\makecell[l]{\textbf{Zero-shot}}} 
& GPT-4o & 5.2 & 12.4 & 31.5 & 2.4 & 2.4 & 6.8 & 14.2 & 37.8 & 3.5 & 3.6\\
& GPT-4o1 & 5.8 & 13.1 & 32.4 & 2.2 & 2.5 & 7.5 & 15.3 & 38.6 & 3.8 & 3.7\\
& Qwen2.5-72B & 3.2 & 10.9 & 30.2 & 2.1 & 2.0 & 5.6 & 12.9 & 36.5 & 3.0 & 3.2\\
& Llama-3.3-70B & 2.0 & 10.6 & 29.9 & 2.0 & 1.9 & 5.3 & 12.6 & 36.2 & 3.0 & 3.1\\
& DeepSeek-R1 & 5.4 & 12.5 & 31.8 & 2.0 & 2.4 & 7.0 & 14.5 & 38.1 & 3.4 & 3.6\\
& Gemini2 & 5.3 & 12.4 & 31.0 & 2.6 & 2.4 & 6.9 & 14.4 & 38.0 & 3.3 & 3.6\\
& Claude3.5 & 5.7 & 13.1 & 32.0 & 2.5 & 2.5 & 7.5 & 14.9 & 38.2 & 3.5 & 3.7\\
\cdashlinelr{1-12}
\multirow{3}{*}{\makecell[l]{\textbf{Retrieval-}\\\textbf{based}}} 
& GPT-4o & 6.2 & 13.8 & 34.6 & 2.5 & 2.6 & 7.8 & 15.8 & 40.9 & 4.0 & 3.8\\
& GPT-4o1 & 6.8 & 14.4 & 35.7 & 2.7 & 2.7 & 8.5 & 16.4 & 42.0 & 4.2 & 3.9\\
& Qwen2.5-72B & 5.2 & 12.8 & 33.3 & 2.3 & 2.2 & 6.6 & 14.8 & 39.6 & 2.9 & 3.4\\
& Llama-3.3-70B & 5.0 & 12.6 & 33.0 & 2.1 & 2.1 & 6.4 & 14.6 & 39.3 & 3.0 & 3.3\\
& DeepSeek-R1 & 6.4 & 14.0 & 34.9 & 2.8 & 2.6 & 8.0 & 16.0 & 41.2 & 4.0 & 3.8\\
& Gemini2 & 6.3 & 13.9 & 34.8 & 2.7 & 2.6 & 7.9 & 15.9 & 41.1 & 3.6 & 3.7\\
& Claude3.5 & 6.6 & 14.0 & 35.8 & 2.6 & 2.5 & 8.3 & 16.5 & 41.9 & 3.8 & 3.8\\
\cdashlinelr{1-12}
\multirow{3}{*}{\makecell[l]{\textbf{Long-}\\\textbf{context}}} 
& Qwen2.5-14B-1M & 6.4 & 14.0 & 35.8 & 2.7 & 2.7 & 8.0 & 16.0 & 42.1 & 3.4 & 3.5\\
& Gemini2 & 7.3 & 15.2 & 36.9 & 2.8 & 2.8 & 8.9 & 17.4 & 43.5 & 4.1 & 4.0\\
& Claude3.5 & 7.7 & 15.3 & 37.6 & 2.8 & 2.9 & 9.3 & 17.3 & 43.9 & 4.3 & 4.1\\
\cdashlinelr{1-12}
\multirow{3}{*}{\makecell[l]{\textbf{\method}}} 
& GPT-4o & 7.8 & 15.4 & 38.9 & 3.0 & 3.0 & 9.4 & 17.4 & 45.2 & 4.5 & 4.2\\
& GPT-4o1 & \textbf{8.6} & \textbf{16.1} & 40.0 & \textbf{3.1} & 3.1 & \textbf{10.1} & \textbf{18.1} & \textbf{46.3} & \textbf{4.6} & \textbf{4.4}\\
& Qwen2.5-72B & 6.8 & 14.4 & 37.6 & 2.9 & 2.8 & 8.4 & 16.4 & 43.9 & 4.2 & 4.0\\
& Llama-3.3-70B & 6.6 & 14.2 & 37.3 & 2.9 & 2.7 & 8.2 & 16.2 & 43.6 & 4.0 & 3.9\\
& DeepSeek-R1 & 7.9 & 15.5 & 39.2 & 3.1 & 3.0 & \textbf{10.1} & 17.5 & 45.5 & 4.0 & 4.2\\
& Gemini2 & 7.8 & 15.4 & 39.1 & 3.0 & 3.0 & 9.4 & 17.4 & 45.4 & 4.1 & 4.2\\
& Claude3.5 & \textbf{8.6} & 16.1 & \textbf{40.2} & 3.0 & \textbf{3.2} & \textbf{10.1} & 17.6 & 45.9 & 4.0 & \textbf{4.4}\\
\bottomrule
\end{tabular}
\caption{Results of different methods and models on \bench. We evaluate both Gold Task (reproducing original thoughts) and Silver Task (generating plausible thoughts) using BLEU, ROUGE-L, NLI scores, and both LLM and Human evaluations. The best scores are \textbf{bolded}.}
\label{tab:thought_generation}
\end{table*}
\subsection{Base Language Models}

After obtaining character profiles, we test multiple LLMs as base models for RPLAs. For long-context approaches, due to length constraints, we evaluate Claude3.5~\citep{anthropic2024claude3}, Gemini2~\citep{geminiteam2024gemini}, and Qwen2.5-14B-1M~\citep{qwen2025qwen25technicalreport}. For other methods, we also test GPT-4o~\citep{openai2023gpt4}, GPT-4o1, Qwen2.5-72B, Llama-3.3-70B~\citep{grattafiori2024llama3herdmodels}, and DeepSeek-R1~\citep{deepseekai2025deepseekr1incentivizingreasoningcapability}. For DeepSeek-R1, we include its chain-of-thought process in our generated thought data. For all these models, we adopt the official instruction formats where available. \footnote{The versions in this paper are \texttt{gpt-4o-2024-11-20, o1-2024-12-17}, \texttt{claude-3-5-sonnet-20241022, gemini-} 
\texttt{2.0-flash, Llama-3.3-70B-Instruct,DeepSeek-R1,}
\texttt{Qwen2.5-72B-Instruct, Qwen2.5-14B-Instruct-1M.}
}

\subsection{Downstream RPLA Tasks}
To validate the effectiveness of high-quality character thought data, we evaluate models on different downstream role-playing tasks. 
We compare two settings: models directly performing role-playing tasks without generating thought data, and models completing tasks after generating thoughts through different methods. 
We conduct experiments on three high-quality role-playing benchmarks: 1) \textit{LifeChoice}, which tests models' ability to reproduce characters' key decisions from the original book; 2) \textit{CROSS-MR}, which evaluates models' ability to generate character behavior motivations; and 3) \textit{RoleEval}, which assesses basic role-playing abilities such as tone and knowledge.
The accuracy of multiple-choice questions serves as the indicator for all these three benchmarks, which also provide the complete memories of the characters. We have tested all the characters in them. 


\section{Experiment Results}
\label{sec:results}

In our experiments, we aim to answer two research questions:
\begin{inparaenum}[\it RQ1)]
\item Can LLMs generate high-quality character thought data?
\item Does character thought data improve role-playing performance?
\end{inparaenum}

\subsection{\textit{Can LLMs generate high-quality character thought data?}}
Table~\ref{tab:thought_generation} shows the experimental results for character inner thought reasoning. The results indicate three main findings. First, generating character thoughts is challenging for LLMs, with models showing moderate performance across multiple evaluation metrics. Second, scores on the Gold Set are generally lower than the Silver Set, as precisely reproducing original thought descriptions is more demanding than generating plausible thought processes. Third, \method-based methods achieve the best performance, followed by Long-context models, demonstrating the importance of accurate memory retrieval for character thought generation.

\paragraph{Method Comparison}
From the method perspective, we analyze different profiling approaches. Zero-shot profiling shows the lowest performance as it relies only on basic character descriptions. While retrieval-based profiling improves performance by accessing character-related memories, it often fails to retrieve the most relevant information. Long-context profiling achieves better results by processing more character information, but suffers from attention dispersion across the extended context. \method addresses these limitations by combining selective memory retrieval with structured reasoning, leading to the best performance among all approaches.

\paragraph{Model Comparison}
From the model perspective, GPT-4o1 and Claude3.5 show balanced performance across all metrics. Models with strong long-text processing capabilities, such as Claude3.5 and Gemini2, perform exceptionally well in Long-context settings. Reasoning-focused models like DeepSeek-R1, despite their chain-of-thought capabilities, do not necessarily excel in this task. While these models typically perform well in mathematical and coding tasks where knowledge is stored in model parameters, character thought generation relies more on accurately capturing scenario-relevant memories than complex reasoning.

\paragraph{Human Analysis}We invite three literature experts to analyze 100 randomly sampled character thoughts from each generation method. Each expert has over 10 years of experience in literary analysis and is familiar with the source material. Their analysis reveals several patterns:
First, compared to references, model-generated thoughts show stronger logical connections but less emotional complexity. Models perform better at expressing explicit emotional states (e.g., anger, fear) than implicit ones (e.g., conflicted loyalty, suppressed guilt).
Second, models consistently maintain character voice and vocabulary preferences, likely influenced by the tendencies in most role-playing task data.
Third, \method-generated thoughts demonstrate stronger narrative coherence, incorporating relevant past events that human readers might overlook. 
This validates the potential for models to perform deeper character understanding tasks, such as character biography generation and behavior analysis.
Finally, the experts conclude that while model-generated character thoughts provide valuable insights for literary character analysis, challenges remain in capturing the full depth of character psychological complexity.
\begin{table}[t]
    \begin{tabular}{ccccc}
    \toprule
    \multicolumn{1}{c}{Models} & 
    \multicolumn{1}{c}{\begin{tabular}[c]{@{}c@{}}\textit{Life}\\\textit{Choice}\end{tabular}} & 
    \multicolumn{1}{c}{\begin{tabular}[c]{@{}c@{}}\textit{CROSS}\\\textit{MR}\end{tabular}} & 
    \multicolumn{1}{c}{\begin{tabular}[c]{@{}c@{}}\textit{Role}\\\textit{Eval}\end{tabular}} \\
    \midrule
    \rowcolor[rgb]{ .949,  .953,  .961} \multicolumn{4}{c}{\textit{Without Thought Generation}} \\
    GPT-4o      & 55.17 & 57.75 & 73.38 \\
    GPT-4o1     & 58.15 & 63.03 & 76.85 \\
    Deepseek-R1 & 54.85 & 60.12 & 74.32 \\
    Gemini2     & 55.03 & 60.42 & 72.01 \\
    Claude3.5   & 58.09 & 65.88 & 76.24 \\
    \midrule
    \rowcolor[rgb]{ .949,  .953,  .961} \multicolumn{4}{c}{\textit{Zero-shot Thought Generation}} \\
    GPT-4o      & 58.32 & 60.85 & 75.52 \\
    GPT-4o1     & 61.25 & 66.18 & 78.93 \\
    Deepseek-R1 & 57.90 & 63.22 & 76.45 \\
    Gemini2     & 58.13 & 63.52 & 74.15 \\
    Claude3.5   & 61.19 & 68.98 & 78.34 \\
    \midrule
    \rowcolor[rgb]{ .949,  .953,  .961} \multicolumn{4}{c}{\textit{Retrieval-based Thought Generation}} \\
    GPT-4o      & 59.89 & 62.45 & 77.08 \\
    GPT-4o1     & 62.82 & 67.78 & 80.49 \\
    Deepseek-R1 & 59.47 & 64.82 & 78.02 \\
    Gemini2     & 59.70 & 65.12 & 75.71 \\
    Claude3.5   & 62.76 & 70.58 & 79.90 \\
    \midrule
    \rowcolor[rgb]{ .949,  .953,  .961} \multicolumn{4}{c}{\textit{Long-context Thought Generation}} \\
    Gemini2     & 60.42 & 66.14 & 76.08 \\
    Claude3.5   & 63.08 & 70.58 & 79.94 \\
    \midrule
    \rowcolor[rgb]{ .949,  .953,  .961} \multicolumn{4}{c}{\textit{MIRROR Thought Generation}} \\
    GPT-4o      & 61.47 & 67.05 & 79.68 \\
    GPT-4o1     & 62.85 & 69.33 & \textbf{82.15} \\
    Deepseek-R1 & 61.15 & 66.42 & 79.62 \\
    Gemini2     & 61.33 & 66.72 & 77.31 \\
    Claude3.5   & \textbf{64.39} & \textbf{72.18} & 80.54 \\
    \bottomrule
    \end{tabular}
    \caption{Performance comparison on downstream RPLA tasks with different thought generation methods and models.}
    \label{tab:downstream_tasks}
\end{table}

\begin{figure}[t]
    \centering    \includegraphics[width=\linewidth]{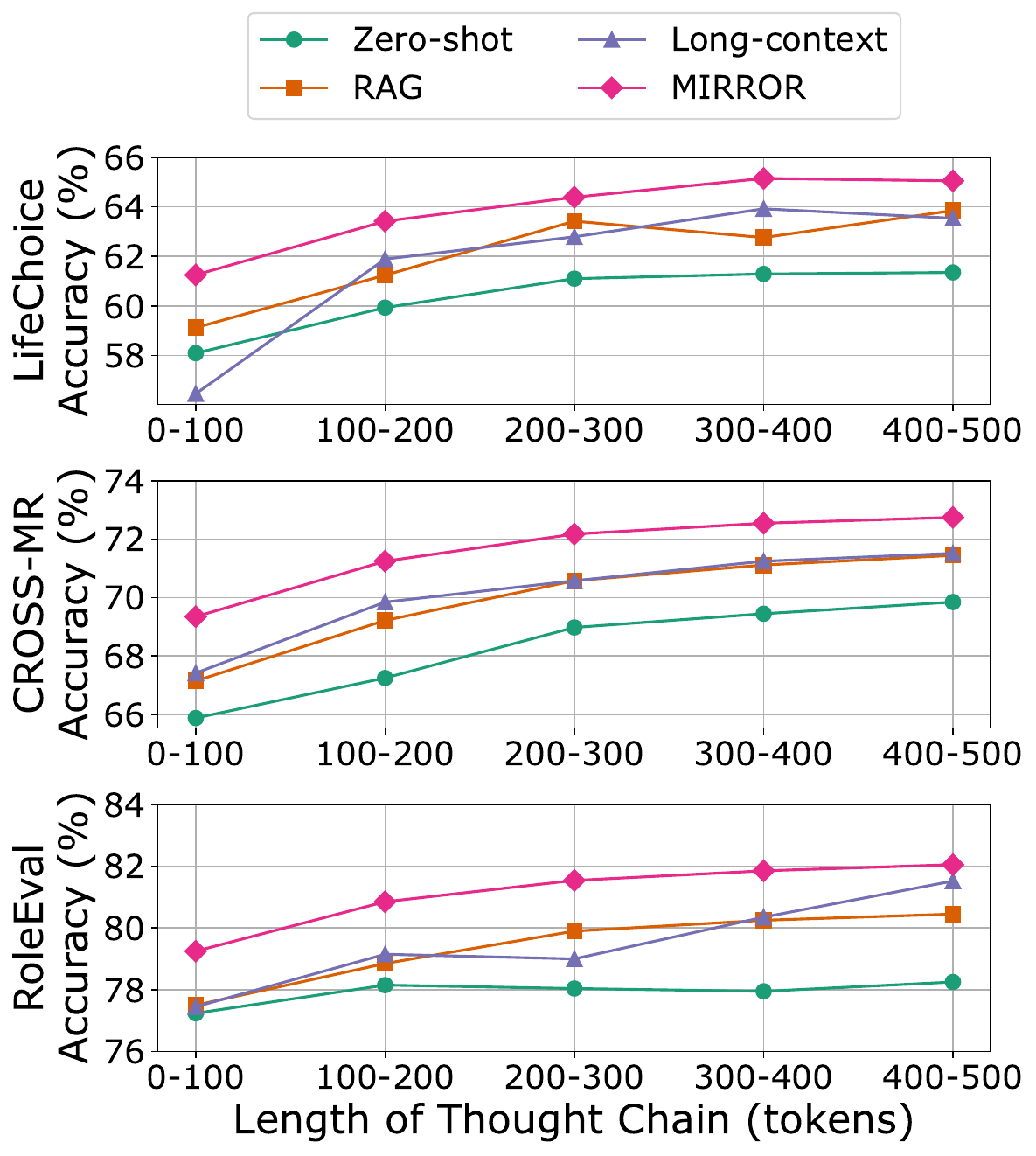}
    \caption{Performance comparison of different thought lengths on downstream tasks. The x-axis represents the token length range of generated thoughts.}
    \label{fig:length}
\end{figure}

\subsection{\textit{Does character thought data improve role-playing performance?}}
Table~\ref{tab:downstream_tasks} demonstrates that character thought data improves performance across all downstream role-playing tasks. For basic tasks like knowledge recall and tone consistency, we observe moderate improvements. For complex tasks such as decision-making and motivation analysis, the improvements are more significant.
The quality of thought data correlates with downstream task performance. High-quality thoughts generated by MIRROR show larger improvements compared to simple zero-shot thoughts, validating the importance of optimizing thought generation. Additionally, these thought processes provide clear explanation chains for model behaviors, making role-playing systems more interpretable and debuggable.

\paragraph{Impact of Thought Length}We analyze how the length of thought affects downstream task performance using Claude-3.5 as the base model. As shown in Figure \ref{fig:length} , longer thought generally lead to better performance across all methods, with several key findings:
First, Performance improvements plateau after 300 tokens, with minimal gains when extending to 500 tokens, suggesting an optimal length range for thought generation.
Second, the impact of thought length varies by task type. Decision-making tasks benefit more from longer thought than knowledge-based tasks, indicating that complex reasoning requires more detailed thought processes.
Finally, the relationship between thought length and task performance depends on the method used. MIRROR maintains high performance even with shorter thought (200 tokens), while other methods require longer (400+ tokens) to achieve similar results. This suggests that MIRROR's memory retrieval mechanism helps create more efficient thought processes.

\begin{table}[t]
    \centering  
    \small
    \begin{tabular}{lcc}
    \toprule
    \rowcolor[gray]{0.95} \multicolumn{3}{c}{\textit{\bench}} \\
    \midrule
    \textbf{Ablation Setting} & \textbf{Gold Set} & \textbf{Silver Set}\\
    \midrule
    Full Model & \underline{3.2} & \underline{4.4}\\
    w/o Memory & 2.4 (-0.8) & 3.4 (-1.0)\\
    w/o ToM & 2.7 (-0.5) & 3.6 (-0.8)\\
    w/o Summary & 3.0 (-0.2) & 4.2 (-0.2)\\
    \midrule
    \rowcolor[gray]{0.95} \multicolumn{3}{c}
    {\textit{Downstream Tasks}} \\
    \midrule
    & \textbf{LifeChoice} & \textbf{CROSS-MR}\\
    \midrule
    Full Model & \underline{64.39} & \underline{72.18}\\
    w/o Memory & 55.29 (-9.1) & 64.88 (-7.3)\\
    w/o ToM & 60.59 (-3.8) & 68.28 (-3.9)\\
    w/o Summary & 62.49 (-1.9) & 69.68 (-2.5)\\
    \bottomrule
    \end{tabular}
    \caption{Ablation study results on \bench tasks and downstream tasks. Numbers in parentheses show performance changes compared to the full model.}
    \label{tab:ablation_study}
\end{table}

\paragraph{Impact of \method Components}As shown in Table \ref{tab:ablation_study}, the ablation studies on \bench and downstream tasks provide insights into the contribution of each component. Memory retrieval shows the most substantial impact, with its removal leading to significant performance drops (-0.8 on Gold sets, -1.0 on Silver sets), highlighting its crucial role in accurate thought generation. The Theory of Mind module demonstrates moderate effects, causing performance degradation on both Gold (-0.5) and Silver sets (-0.8). While the summary component shows the smallest impact (-0.2 across tasks), it consistently contributes to model performance. These effects become more pronounced in downstream tasks, particularly in decision-making scenarios like LifeChoice (-9.1 without memory retrieval) and CROSS-MR (-7.3 without memory retrieval), indicating that complete thought processes are essential for complex reasoning tasks.
\section{Conclusion}
In this paper, we present the first systematic study on generating inner thought processes for role-playing language agents (RPLAs). We introduce \bench, a benchmark constructed from \textit{A Song of Ice and Fire}, which contains both direct character monologues and expert-analyzed character behaviors. Our proposed method, \method, shows significant improvements in character thought generation by combining memory retrieval, role reasoning, and observation reflection. Our experiments with various LLMs demonstrate that generating complex psychological processes remains a challenging task. High-quality thought generation can both provide deeper character understanding and analysis, while also improving RPLAs' performance across multiple downstream tasks.

\section*{Limitations}
\label{sec:limitation}
While our work presents important advances in understanding and generating character thought processes, several limitations should be noted.
Our benchmark \bench is constructed using a single source material (\textit{A Song of Ice and Fire}). Although this series provides rich character development and psychological descriptions, the thought patterns and decision-making processes might be specific to this particular literary work. The effectiveness of our approach on characters from different genres, cultural backgrounds, or writing styles remains to be explored.
The evaluation of thought processes relies heavily on comparing generated thoughts with references, either from original monologues or expert analyses. This approach assumes that the references accurately represent characters' true thought processes. However, even expert analyses can be subjective, and original monologues might not fully capture the complexity of character psychology.

\section*{Ethics Statement}
\label{sec:Ethics}
\paragraph{Use of Human Annotations}
Our research involves literature experts who provide character analyses and evaluations for the Silver task in our \bench benchmark. These experts are compensated above local minimum wage standards and have provided informed consent for the use of their analyses in our research. We maintain strict privacy protocols to protect their identities and personal information throughout the research process.
\paragraph{Risks}
While our benchmark is constructed from a published literary work, we acknowledge potential risks in character thought generation. The generated thoughts might contain biased or inappropriate content, reflecting both the source material's content and potential biases in language models. Additionally, our method of analyzing character psychology could be misused to manipulate or deceive if applied to real-world scenarios. We emphasize that our research is focused on fictional characters and should not be used for psychological analysis of real individuals.

\bibliography{anthology,custom}
\clearpage
\appendix
\section{Dataset}
\label{sec:appendix_dataset}

\begin{table*}[htbp]
    \small
    \centering
    \resizebox{\linewidth}{!}{
    \begin{tabular}{p{2.1in}|p{2.1in}|p{2.1in}}
    \toprule
    \multicolumn{3}{c}{\textbf{Selected Characters}} \\  \hline
    \textbf{1}. \textit{Tyrion Lannister} & \textbf{2}. \textit{Jon Snow} & \textbf{3}. \textit{Arya Stark}\\ \hline
    \textbf{4}. \textit{Daenerys Targaryen} & \textbf{5}. \textit{Catelyn Stark} & \textbf{6}. \textit{Sansa Stark}\\ \hline
    \textbf{7}. \textit{Bran Stark} & \textbf{8}. \textit{Jaime Lannister} & \textbf{9}. \textit{Eddard Stark}\\ \hline
    \textbf{10}. \textit{Theon Greyjoy} & \textbf{11}. \textit{Davos Seaworth} & \textbf{12}. \textit{Cersei Lannister}\\ \hline
    \textbf{13}. \textit{Samwell Tarly} & \textbf{14}. \textit{Brienne of Tarth} & \textbf{15}. \textit{Victarion Greyjoy}\\ \hline
    \textbf{16}. \textit{Asha Greyjoy} & \textbf{17}. \textit{Arianne Martell} & \textbf{18}. \textit{Barristan Selmy}\\ \hline
    \textbf{19}. \textit{Quentyn Martell} & \textbf{20}. \textit{Jon Connington} & \textbf{21}. \textit{Melisandre}\\ \hline
    \textbf{22}. \textit{Aeron Greyjoy} & & \\
    \bottomrule
    \end{tabular}}
    \caption{The selected characters from A Song of Ice and Fire series. Characters 1-22 are used in the Gold Task, while characters 1-18 are used in the Silver Task.}
    \label{tab:selected_characters}
\end{table*}

\subsection{Dataset Statistics}
As shown in Section \ref{sec:bench_dataset}, for the Gold Set, we have 405 data points from 22 characters, and for the Silver Set, we have 211 data points from 18 characters. Table~\ref{tab:selected_characters} lists all the selected characters.

\subsection{Data Usage and Permissions}
For the Silver Sets, we collected data from three major fan websites dedicated to analyzing \textit{A Song of Ice and Fire}. We obtained explicit permission from the website administrators and content creators for academic use. The data collection process strictly followed the websites' terms of service and data usage policies. All content creators were informed about the research purpose and agreed to have their analyses included in our dataset. Additionally, we ensured that our data collection and usage comply with fair use guidelines for academic research.

\section{Prompt}
\label{sec:appendix_prompt}

\begin{table*}[t]
    \centering
    \small
    \begin{tabular}{l}
    \toprule
    \rowcolor[gray]{0.95}\multicolumn{1}{c}{\textbf{Prompt I: Character Identification}} \\
    \makecell[l]{
    Please analyze the important characters in the following text:\\
    <text>\\\\
    \color[rgb]{0,0.39,0}{\# Output Format:}\\
    Return the character list in JSON format as follows:\\
    \{"characters": ["character1", "character2", ...]\}
    }\\
    \bottomrule
    \end{tabular}
    \caption{Prompt templates for character identification in Gold Task.}
    \label{tab:prompt_char_identification}
\end{table*}
\begin{table*}[t]
    \centering
    \small
    \begin{tabular}{l}
    \toprule
    \rowcolor[gray]{0.95}\multicolumn{1}{c}{\textbf{Prompt II: Thought Extraction}} \\
    \makecell[l]{
    Extract the thoughts of character <character> from the following text.\\\\
    \color[rgb]{0,0.39,0}{\# Requirements:}\\
    1. Only return high-quality thought segments that reflect the character's internal mental process\\
    2. Thoughts should be coherent and contain at least two sentences\\
    3. Thoughts must be directly quoted from the original text, without any modification\\
    4. Thoughts should be purely internal monologues, not:\\
    \quad - Spoken dialogue\\
    \quad - Physical actions\\
    \quad - Narrative descriptions\\
    \quad - External observations\\
    \color[rgb]{0,0.39,0}{\# Output Format:}\\
    \{\\
    \quad "ta\_pairs": [\\
    \quad \quad \{\\
    \quad \quad \quad "character": "character\_name",\\
    \quad \quad \quad "reason": "explanation for selecting this thought segment",\\
    \quad \quad \quad "thought": "extracted thought content",\\
    \quad \quad \quad "raw\_text": "original text segment"\\
    \quad \quad \}\\
    \quad ]\\
    \}\\\\
    \color[rgb]{0,0.39,0}{\# Input:}\\
    <text>
    }\\
    \bottomrule
    \end{tabular}
    \caption{Prompt templates for thought extraction in Gold Task.}
    \label{tab:prompt_thought_extraction}
\end{table*}
\begin{table*}[t]
    \centering
    \small
    \begin{tabularx}{\textwidth}{X}
    \toprule
    \rowcolor[gray]{0.95}\multicolumn{1}{c}{\textbf{Prompt III: Thought Generation}} \\
    \makecell[l]{
    Your task is to generate the masked thoughts of a character in the given scenario.\\\\
    \color[rgb]{0,0.39,0}{\# Inputs:}\\
    1. Character Profile:\\
    <profile>\\\\
    2. Scenario Context:\\
    <context>\\\\
    3. Masked Thought Location:\\
    <text with [MASK] indicating the thought position>\\\\
    \color[rgb]{0,0.39,0}{\# Requirements:}\\
    1. Generate thoughts that are consistent with the character's personality and background\\
    2. Ensure the thoughts fit naturally into the given context\\
    3. Match the emotional state and decision-making process implied by the scenario\\
    4. Maintain the character's perspective and knowledge at that specific moment\\\\
    \color[rgb]{0,0.39,0}{\# Output:}\\
    Generate the content that should replace [MASK], representing the character's inner thoughts.
    }\\
    \bottomrule
    \end{tabularx}
    \caption{Prompt templates for thought generation in Gold Task.}
    \label{tab:prompt_thought_generation}
\end{table*}

\subsection{Prompts for Gold Set}
For the Gold Set, we design three types of prompts to systematically extract and generate character thoughts. As shown in Table \ref{tab:prompt_char_identification}, the first prompt identifies key characters in POV chapters, focusing on characters who play significant roles in the current scenario. The second prompt, presented in Table \ref{tab:prompt_thought_extraction}, locates high-quality thought segments for these key characters, specifically requiring coherent internal monologues that reveal decision-making processes and emotional reactions. In Table \ref{tab:prompt_thought_generation}, we present the third prompt designed for thought generation, where we provide the character's profile and scenario context, asking the model to generate thoughts that align with the character's personality and fit naturally into the given context. All prompts are carefully designed to ensure consistent evaluation and maintain the quality of extracted and generated thoughts across experiments.

\begin{table*}[t]
    \centering
    \small
    \begin{tabularx}{\textwidth}{X}
    \toprule
    \rowcolor[gray]{0.95}\multicolumn{1}{c}{\textbf{Prompt IV: Motivation Analysis}} \\
    \makecell[l]{
    Analyze the following fan-written character analysis article and extract structured information.\\\\
    \color[rgb]{0,0.39,0}{\# Input:}\\
    <article>\\\\
    \color[rgb]{0,0.39,0}{\# Requirements:}\\
    1. Identify the main character being analyzed\\
    2. Extract specific behaviors or decisions mentioned in the article\\
    3. For each behavior, identify the detailed motivations and reasoning behind it\\\\
    \color[rgb]{0,0.39,0}{\# Output Format:}\\
    \{\\
    \quad "character": "character\_name",\\
    \quad "behaviors": [\\
    \quad \quad \{\\
    \quad \quad \quad "action": "specific behavior or decision",\\
    \quad \quad \quad "motivations": [\\
    \quad \quad \quad \quad "detailed reason 1",\\
    \quad \quad \quad \quad "detailed reason 2",\\
    \quad \quad \quad \quad ...\\
    \quad \quad \quad ]\\
    \quad \quad \}\\
    \quad ]\\
    \}
    }\\
    \bottomrule
    \end{tabularx}
    \caption{Prompt templates for motivation analysis in Silver Task.}
    \label{tab:prompt_silver_motivation}
\end{table*}
\begin{table*}[t]
    \centering
    \small
    \begin{tabularx}{\textwidth}{X}
    \toprule
    \rowcolor[gray]{0.95}\multicolumn{1}{c}{\textbf{Prompt V: Memory Recall}} \\
    \makecell[l]{
    As the character, recall all memories that are relevant to the current scenario.\\\\
    \color[rgb]{0,0.39,0}{\# Inputs:}\\
    1. Character Profile:\\
    <profile>\\\\
    2. Current Scenario:\\
    <scenario>\\\\
    \color[rgb]{0,0.39,0}{\# Requirements:}\\
    1. Think as the character\\
    2. Recall any past experiences, events, or knowledge related to this scenario\\
    3. Consider both direct and indirect connections\\\\
    \color[rgb]{0,0.39,0}{\# Output Format:}\\
    \{\\
    \quad "memories": [\\
    \quad \quad \{"memory": "description of the memory",\\
    \quad \quad "relevance": "why this memory is relevant to current scenario"\}\\
    \quad ]\\
    \}
    }\\
    \bottomrule
    \end{tabularx}
    \caption{Prompt template for memory recall in MIRROR.}
    \label{tab:prompt_memory_recall}
\end{table*}
\begin{table*}[t]
    \centering
    \small
    \begin{tabularx}{\textwidth}{X}
    \toprule
    \rowcolor[gray]{0.95}\multicolumn{1}{c}{\textbf{Prompt VI: Theory of Mind Thinking}} \\
    \makecell[l]{
    As the character, analyze how other characters, groups, or environments might react to your potential actions in this scenario.\\\\
    \color[rgb]{0,0.39,0}{\# Inputs:}\\
    1. Character Profile:\\
    <profile>\\\\
    2. Current Scenario:\\
    <scenario>\\\\
    \color[rgb]{0,0.39,0}{\# Requirements:}\\
    1. Identify relevant objects (characters, groups, environments)\\
    2. Predict their possible reactions\\
    3. Consider their perspectives and motivations\\\\
    \color[rgb]{0,0.39,0}{\# Output Format:}\\
    \{\\
    \quad "objects": [\\
    \quad \quad \{"object": "name or description",\\
    \quad \quad "relationship": "relationship with the character",\\
    \quad \quad "predicted\_reaction": "how they might react and why"\}\\
    \quad ]\\
    \}
    }\\
    \bottomrule
    \end{tabularx}
    \caption{Prompt template for Theory of Mind thinking in MIRROR.}
    \label{tab:prompt_tom}
\end{table*}
\begin{table*}[t]
    \centering
    \small
    \begin{tabularx}{\textwidth}{X}
    \toprule
    \rowcolor[gray]{0.95}\multicolumn{1}{c}{\textbf{Prompt VII: Reflection \& Summarization}} \\
    \makecell[l]{
    As the character, reflect on the recalled memories and predicted reactions to generate your inner thoughts.\\\\
    \color[rgb]{0,0.39,0}{\# Inputs:}\\
    1. Character Profile:\\
    <profile>\\\\
    2. Current Scenario:\\
    <scenario>\\\\
    3. Recalled Memories:\\
    <memories>\\\\
    4. Theory of Mind Analysis:\\
    <predictions>\\\\
    \color[rgb]{0,0.39,0}{\# Requirements:}\\
    1. Remove any memories or predictions that are not directly relevant\\
    2. Filter relevant information from remaining content\\
    3. Organize thoughts in a coherent way\\
    4. Ensure the thought process aligns with character's personality\\\\
    \color[rgb]{0,0.39,0}{\# Output Format:}\\
    \{\\
    \quad "character": "character name",\\
    \quad "inner\_thoughts": "character's organized thought process" \\
    \}
    }\\
    \bottomrule
    \end{tabularx}
    \caption{Prompt template for reflection and summarization in MIRROR.}
    \label{tab:prompt_reflection}
\end{table*}

\begin{table*}[t]
    \centering
    \small
    \begin{tabularx}{\textwidth}{X}
    \toprule
    \rowcolor[gray]{0.95}\multicolumn{1}{c}{\textbf{Prompt VIII: Thought Point Location}} \\
    \makecell[l]{
    Locate the specific point in the chapter where the character's motivation might have manifested as internal thoughts.\\\\
    \color[rgb]{0,0.39,0}{\# Inputs:}\\
    1. Character Motivation:\\
    <motivation>\\\\
    2. Chapter Content:\\
    <chapter>\\\\
    \color[rgb]{0,0.39,0}{\# Requirements:}\\
    1. Find the most appropriate point where the character might have had these thoughts\\
    2. The point should be before the actual behavior or decision\\
    3. The location should have sufficient context for understanding the situation\\\\
    \color[rgb]{0,0.39,0}{\# Output Format:}\\
    \{\\
    \quad "thought\_point": \{\\
    \quad \quad "location": "text segment before the thought point",\\
    \quad \quad "reason": "explanation for choosing this point"\\
    \quad \}\\
    \}
    }\\
    \bottomrule
    \end{tabularx}
    \caption{Prompt templates for thought point location in Silver Task.}
    \label{tab:prompt_silver_location}
\end{table*}
\begin{table*}[t]
    \centering
    \small
    \begin{tabularx}{\textwidth}{X}
    \toprule
    \rowcolor[gray]{0.95}\multicolumn{1}{c}{\textbf{Prompt IX: Character Thought Generation}} \\
    \makecell[l]{
    You are the character described in the profile. Generate your detailed thoughts at this specific moment.\\\\
    \color[rgb]{0,0.39,0}{\# Inputs:}\\
    1. Character Profile:\\
    <profile>\\\\
    2. Current Scenario:\\
    <context>\\\\
    \color[rgb]{0,0.39,0}{\# Requirements:}\\
    1. Generate detailed internal thoughts from the character's perspective\\
    2. Ensure consistency with the character's personality and background\\
    3. Consider only information available to the character at this moment\\\\
    \color[rgb]{0,0.39,0}{\# Output:}\\
    Write a detailed inner monologue expressing your thoughts at this moment.
    }\\
    \bottomrule
    \end{tabularx}
    \caption{Prompt templates for character thought generation in Silver Task.}
    \label{tab:prompt_silver_generation}
\end{table*}
\begin{table*}[t]
    \centering
    \small
    \begin{tabular}{l}
    \toprule
    \rowcolor[gray]{0.95}\multicolumn{1}{c}{\textbf{Prompt X: Gold Set Evaluation}} \\
    \makecell[l]{
    You are evaluating the quality of generated character thoughts compared to the reference thoughts.\\\\
    \color[rgb]{0,0.39,0}{\# Inputs:}\\
    1. Reference Thought:\\
    <reference>\\\\
    2. Generated Thought:\\
    <generated>\\\\
    \color[rgb]{0,0.39,0}{\# Scoring Criteria:}\\
    5 points:\\
    - Contains ALL elements from the reference thought\\
    - Provides reasonable additional context or elaboration\\
    - Maintains perfect character voice and perspective\\
    - Shows deep understanding of the character's state\\
    - Additional content logically connects to the reference\\\\
    4 points:\\
    - Contains ALL elements from the reference thought\\
    - Provides some additional context\\
    - Maintains character voice\\
    - Shows good understanding\\
    - No contradictions or inconsistencies\\\\
    3 points:\\
    - Contains MOST elements from the reference thought\\
    - Limited or no additional context\\
    - Generally maintains character voice\\
    - Shows basic understanding\\
    - May miss minor elements\\\\
    2 points:\\
    - Contains SOME elements from the reference thought\\
    - Missing major elements\\
    - Inconsistent character voice\\
    - Shows limited understanding\\
    - May contain minor contradictions\\\\
    1 point:\\
    - Missing MOST reference elements\\
    - Wrong character voice\\
    - Shows no understanding\\
    - Contains major contradictions\\
    - Completely different direction\\\\
    \color[rgb]{0,0.39,0}{\# Output:}\\
    Provide a score (1-5) with a brief explanation of your rating.
    }\\
    \bottomrule
    \end{tabular}
    \caption{Evaluation prompt for Gold Set.}
    \label{tab:prompt_eval_gold}
\end{table*}
\begin{table*}[t]
    \centering
    \small
    \begin{tabular}{l}
    \toprule
    \rowcolor[gray]{0.95}\multicolumn{1}{c}{\textbf{Prompt XI: Silver Set Evaluation}} \\
    \makecell[l]{
    You are evaluating the quality of generated character thoughts based on character motivations and context.\\\\
    \color[rgb]{0,0.39,0}{\# Inputs:}\\
    1. Character Profile:\\
    <profile>\\\\
    2. Scenario Context:\\
    <context>\\\\
    3. Generated Thought:\\
    <generated>\\\\
    \color[rgb]{0,0.39,0}{\# Scoring Criteria:}\\
    5 points:\\
    - Perfect alignment with known character motivations\\
    - Rich, multi-layered reasoning process\\
    - Deep consideration of current context\\
    - Consistent with character's knowledge at this point\\
    - Natural connection to subsequent actions\\\\
    4 points:\\
    - Strong alignment with character motivations\\
    - Clear reasoning process\\
    - Good consideration of context\\
    - Consistent with character's knowledge\\
    - Logical connection to actions\\\\
    3 points:\\
    - Basic alignment with character motivations\\
    - Simple but logical reasoning\\
    - Some consideration of context\\
    - Generally consistent with knowledge\\
    - Basic connection to actions\\\\
    2 points:\\
    - Weak alignment with character motivations\\
    - Unclear or illogical reasoning\\
    - Limited context consideration\\
    - Some knowledge inconsistencies\\
    - Weak connection to actions\\\\
    1 point:\\
    - No alignment with character motivations\\
    - No clear reasoning\\
    - Ignores context\\
    - Major knowledge inconsistencies\\
    - No connection to actions\\\\
    \color[rgb]{0,0.39,0}{\# Output:}\\
    Provide a score (1-5) with a brief explanation of your rating.
    }\\
    \bottomrule
    \end{tabular}
    \caption{Evaluation prompt for Silver Set.}
    \label{tab:prompt_eval_silver}
\end{table*}
\subsection{Prompts for Silver Set}
For the Silver Set, we design three prompts to process fan-based character analyses and generate novel thoughts. As shown in Table \ref{tab:prompt_silver_motivation}, the first prompt analyzes character-focused articles from fan websites, extracting structured information including the character name, their specific behaviors, and detailed motivations behind these behaviors. The second prompt, presented in Table \ref{tab:prompt_silver_location}, takes these extracted motivations along with the corresponding chapter content to locate specific points where these thoughts might have occurred. In Table \ref{tab:prompt_silver_generation}, we present the third prompt that focuses on thought generation, where we provide the scenario context up to the identified point and the character's profile, asking the model to generate plausible thinking processes from the character's perspective. These prompts work together to ensure that the generated thoughts are both consistent with fan-sourced character analyses and naturally fit into the story context.

\subsection{Prompts for \method}
\label{sec:appendix_mirror}
To implement our \method approach, we design three carefully crafted prompts that guide the model through the chain-of-thought process. As shown in Table \ref{tab:prompt_memory_recall}, the first prompt focuses on memory recall, where we ask the model to retrieve relevant memories from the character's perspective based on the current scenario. The second prompt, presented in Table \ref{tab:prompt_tom}, implements Theory of Mind thinking by guiding the model to analyze potential reactions from other characters, groups, or environments that might influence the character's thoughts. In Table \ref{tab:prompt_reflection}, we present the third prompt for reflection and summarization, which helps the model filter out irrelevant information and organize the remaining content into a coherent thought process that aligns with the character's personality. These prompts work together to ensure that the generated thoughts are grounded in the character's memories, socially aware through perspective-taking, and coherently structured through careful reflection.

\section{Human Filtering and Evaluation
}
\subsection{Data Filtering Criteria}
To ensure data quality, we establish comprehensive filtering criteria for both sets:
\begin{itemize}
    \item \textbf{Gold Set Filtering:}
    \begin{itemize}
        \item Remove segments that contain:
            \begin{itemize}
                \item Narrative descriptions of character actions
                \item Dialogue or spoken words
                \item Physical sensations or reactions
            \end{itemize}
        \item Only retain segments that:
            \begin{itemize}
                \item Represent pure internal monologue
                \item Show complete thought processes
            \end{itemize}
    \end{itemize}
    
    \item \textbf{Silver Set Filtering:}
    \begin{itemize}
        \item Remove cases where:
            \begin{itemize}
                \item The extracted motivation lacks clear reasoning
                \item The located thought point is after the action
                \item Multiple characters' thoughts are mixed
            \end{itemize}
        \item Only retain cases that:
            \begin{itemize}
                \item Have well-documented character motivations
                \item Provide clear decision-making context
                \item Maintain consistency with character profiles
            \end{itemize}
    \end{itemize}
\end{itemize}

Through this filtering process, we removed approximately 63.4\% of the automatically extracted data, ensuring that our benchmark contains only high-quality examples suitable for evaluation.

\subsection{Evaluation Guidelines}
\label{sec:appendix_eval}
We establish comprehensive scoring criteria for both human annotators and LLM evaluators through carefully designed prompts. For the Gold Set (Table \ref{tab:prompt_eval_gold}), evaluators compare generated thoughts with reference content from the original text. While containing all reference elements is necessary for high scores, we also evaluate the quality of additional reasoning and maintenance of character voice. For the Silver Set (Table \ref{tab:prompt_eval_silver}), evaluators assess how well the generated thoughts align with the character's established motivations and knowledge state, considering both the reasoning depth and contextual consistency. To ensure evaluation consistency, both human annotators and LLM evaluators follow the same structured prompts and scoring criteria. All the annotators are fans of *A Song of Ice and Fire* and hold a college diploma. For all the annotators, we offer compensation for the tasks at the local minimum - hourly - wage standard.  

\end{document}